\documentclass[11pt,a4paper]{article}
\usepackage[hyperref]{emnlp2020}
\usepackage{times}
\usepackage{latexsym}

\usepackage{microtype}
\usepackage{amsfonts,amssymb}
\usepackage{graphicx}
\usepackage{amsmath}
\usepackage{multirow}
\usepackage{booktabs}
\usepackage{amsfonts}
\usepackage{url}
\usepackage{pgfplots}

\definecolor{color1}{RGB}{57, 197, 200}
\definecolor{color2}{RGB}{182, 163, 200}
\definecolor{color3}{RGB}{94, 178, 237}
\definecolor{color4}{RGB}{253, 185, 132}
\definecolor{color5}{RGB}{214, 123, 129}

\usepackage{subfigure}
\usepackage{latexsym}
\usepackage{arydshln}
\aclfinalcopy 

\title{Reference Language based Unsupervised Neural Machine Translation}

\author{
	Zuchao Li$^{1,2,3}$,
	Hai Zhao$^{1,2,3,}$\thanks{$\ $ Corresponding authors. This paper was partially supported by National Key Research and Development Program of China (No. 2017YFB0304100), Key Projects of National Natural Science Foundation of China (U1836222 and 61733011), Huawei-SJTU Long Term AI Project, Cutting-edge Machine Reading Comprehension and Language Model. Rui Wang was partially supported by JSPS grant-in-aid for early-career scientists (19K20354): ``Unsupervised Neural Machine Translation in Universal Scenarios" and NICT tenure-track researcher startup fund ``Toward Intelligent Machine Translation".},
	Rui Wang$^{4,*}$,
	Masao Utiyama$^{4}$,
	and Eiichiro Sumita$^{4}$
	\\
	$^1$Department of Computer Science and Engineering, Shanghai Jiao Tong University (SJTU)\\
	$^2$Key Laboratory of Shanghai Education Commission for Intelligent Interaction\\
	and Cognitive Engineering, Shanghai Jiao Tong University, Shanghai, China\\
	$^3$MoE Key Lab of Artificial Intelligence, AI Institude, Shanghai Jiao Tong University, China\\
	$^4$National Institute of Information and Communications Technology (NICT), Kyoto, Japan \\
	{\tt charlee@sjtu.edu.cn, zhaohai@cs,sjtu.edu.cn,} \\
	{\tt\{wangrui, mutiyama, eiichiro.sumita\}@nict.go.jp}
}

\date{}

\begin{document}
\maketitle
\begin{abstract}
Exploiting a common language as an auxiliary for better translation has a long tradition in machine translation and lets supervised learning-based machine translation enjoy the enhancement delivered by the well-used pivot language in the absence of a source language to target language parallel corpus. The rise of unsupervised neural machine translation (UNMT) almost completely relieves the parallel corpus curse, though UNMT is still subject to unsatisfactory performance due to the vagueness of the clues available for its core back-translation training. 
Further enriching the idea of pivot translation by extending the use of parallel corpora beyond the source-target paradigm, we propose a new reference language-based framework for UNMT, RUNMT, in which the reference language only shares a parallel corpus with the source, but this corpus still indicates a signal clear enough to help the reconstruction training of UNMT through a proposed reference agreement mechanism.
Experimental results show that our methods improve the quality of UNMT over that of a strong baseline that uses only one auxiliary language, demonstrating the usefulness of the proposed reference language-based UNMT and establishing a good start for the community.

\end{abstract}

\section{Introduction}

Recently, the application of neural machine translation (NMT) \cite{sutskever2014sequence,bahdanau2014neural}  to standard benchmarks has achieved great success \cite{wu2016google,gehring2017convolutional,vaswani2017attention}  because of advances in deep learning and the availability of large-scale parallel corpora; however, the applicability of MT systems is limited because of their reliance on large parallel corpora for the majority of language pairs.  In real-world situations, the majority of language pairs have very little parallel data, although large volumes of monolingual data are available for each language. UNMT removes the dependence on parallel corpora, relying only on monolingual corpora in each language \cite{artetxe2018unsupervised,lample2018unsupervised,lample2018phrase,conneau2019cross,li2019data}.

\begin{figure}[t]
	\centering
	\includegraphics[width=0.48\textwidth]{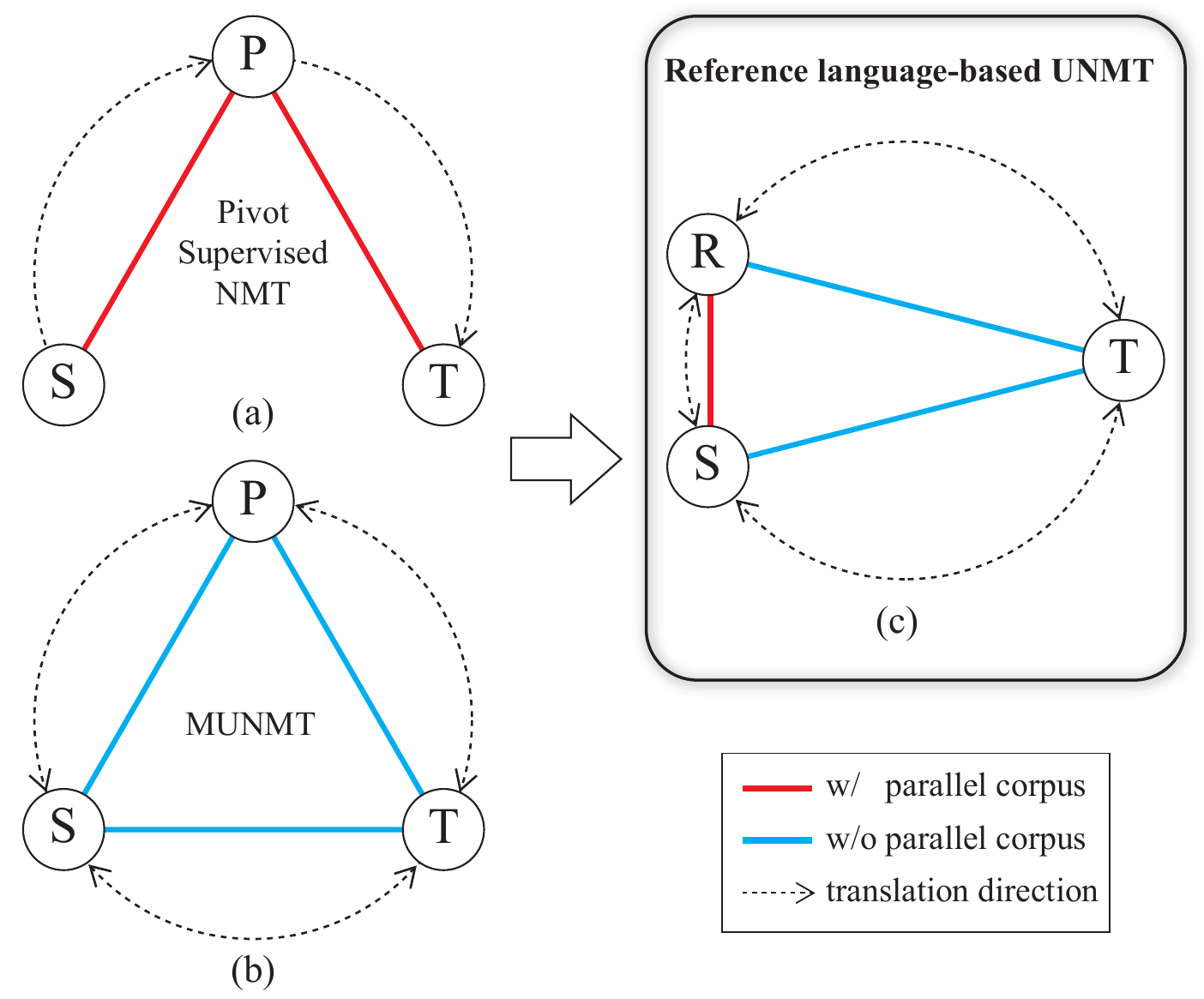}
	\caption{Schemas of (a) pivot supervised NMT, (b) MUNMT, (c) our proposed RUNMT, where $\mathcal{S}$ stands for source language, $\mathcal{T}$ for target language, $\mathcal{P}$ for pivot language in pivot translation, and $\mathcal{R}$ for the reference language in RUNMT.} \label{RUNMT}
\end{figure}

UNMT uses translation symmetry for dual learning in each language direction. Existing UNMT models are mainly built on the encoder--decoder schema. The essence of UNMT is to learn unsupervised cross-lingual word alignment and/or sentence alignment. For unsupervised word alignment, the most popular methods are word embedding mapping \cite{conneau2017word,lample2018unsupervised,sun2019unsupervised}, vocabulary sharing \cite{lample2018phrase}, and language modeling  \cite{conneau2019cross}. Weight sharing can also be adopted in the encoder/decoder, adversarial training, and back-translation (BT) processes for unsupervised sentence alignment.

BT aims to train models using iteratively generated pseudo-parallel data, thus overcoming the lack of cross-language signals. Specifically, monolingual data in the source language is translated to the target language using a source-to-target translation model, and then the pseudo-parallel data (including both the generated and the original data) is used to train the target-to-source translation model, and vice versa.

Unfortunately, as the input sentences in the pseudo-parallel data are generated by unsupervised models, random errors and noise are inevitably introduced, resulting in low-quality parallel data for model training and bad translation performance. 
In addition, when vocabulary sharing UNMT models for two distant languages (that is, very little vocabulary overlap between the source language and target language) are trained with BT,  the unsupervised model may generate the words in the source language instead of in the target language under source-to-target forward translation. 
As a result, although the reconstruction loss is small if the forward translation generation is very similar to the input, the model is not sufficiently optimized because the pseudo-parallel corpus contains very little cross-lingual sentence alignment information.

Multilingualism \cite{edwards2002multilingualism,clyne2017multilingualism} is a powerful fact of communication across speech communities. In multilingualism, an important ``lingua franca'' (or common language) often serves as an aid to cross-group understanding, usually representing the language of a potent and prestigious society with a large number of users. For machine translation, the parallel corpora between languages and some lingua franca are usually more abundant. 
Thus, conventional Pivot Translation (PT) usually leverages a resource-rich language (mainly English) as the pivot to help the low/zero-resource translation (see Appendix \ref{pivot_translation} for a detailed analysis).
Although UNMT no longer requires parallel corpora, this feature is still worth exploring and can be used to enhance current UNMT systems under low- or zero-resource scenarios. In addition, we can further use the transfer learning capabilities of the model to transfer the translation capabilities of languages and lingua francas to any two languages that need to be learned.

In this work, taking the merits of pivot language translation in both supervised NMT and UNMT as shown in Figure \ref{RUNMT}, we propose the reference language-based UNMT framework in which the reference language shares a parallel corpus with only the source language (using only the target language follows a similar pattern). In the framework, we use multilingualism and propose a reference agreement mechanism. Exploiting the accurate alignment clues between source and reference languages, we can more confidently enhance source-target UNMT by taking into account the translation agreement within the source, reference, and target languages. 
Specifically, this previously irrelevant parallel data plays a role in controlling the quality of the pseudo-sentence pairs through a cross-lingual equivalence (translation agreement). The proposed mechanism is orthogonal to the common multilingual transfer learning methods and different from the general pivot translation method.
Empirical results on popular benchmarks and distant languages show that the reference agreement mechanism consistently improves the performance of UNMT systems. In addition, we explore the impact of multilingual information on the basis of our multilingual UNMT baseline and proposed method.

\begin{figure*}[t]
	\centering
	\includegraphics[width=0.98\textwidth]{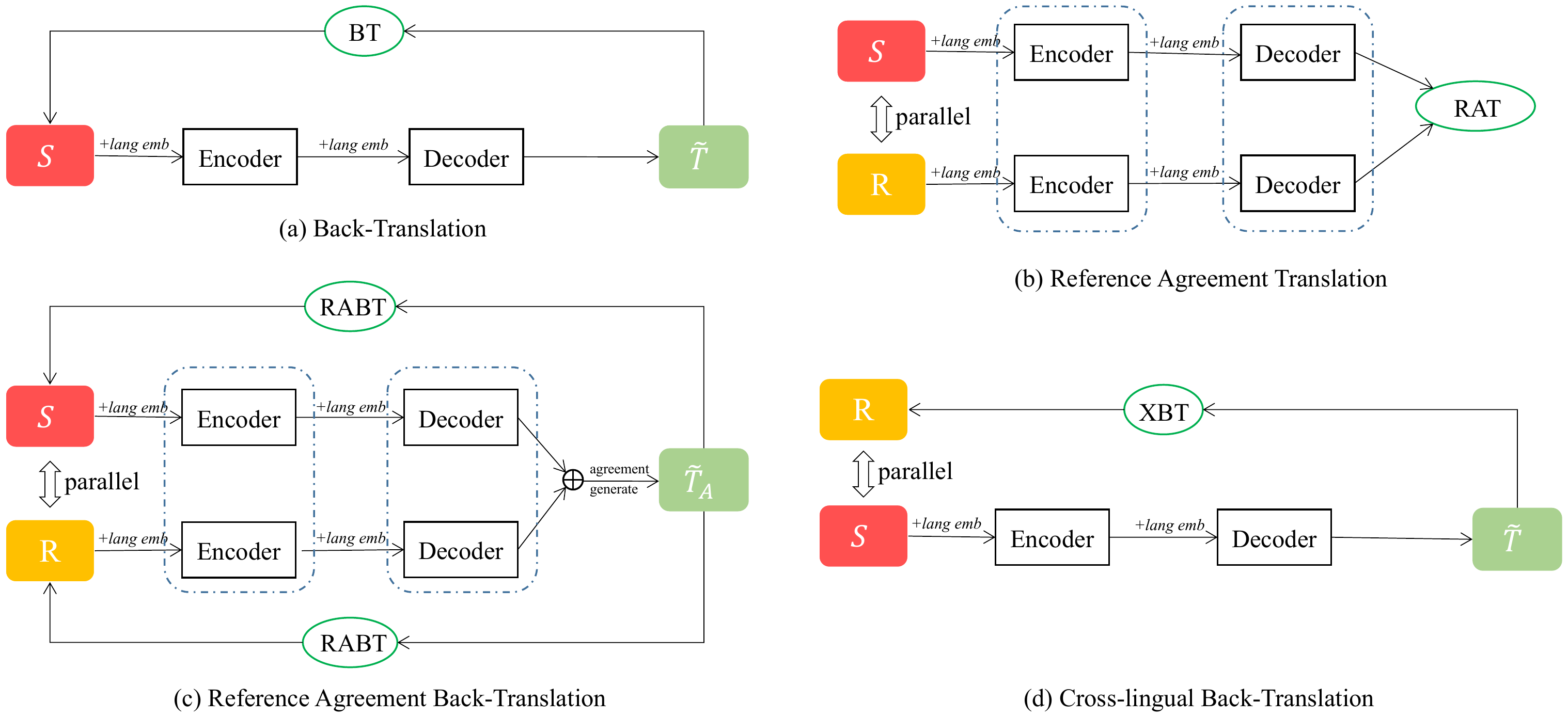}
	\caption{Illustration of (a) Back-Translation (BT), (b) Reference Agreement Translation (RAT), (c) Reference Agreement Back-Translation (RABT), and (d) Cross-lingual Back-Translation (XBT) in UNMT. The green ellipses represents the name of the corresponding process, the arrow pointing to the ellipses represents the input for loss calculation, while pointing out indicates optimization target.} \label{methods}
\end{figure*}

\section{UNMT}\label{unmt}

UNMT is a recently proposed MT paradigm that attempts to achieve the co-growth of MT models in two directions while relying solely on monolingual data and for example, would benefit both English-to-French vs. French-to-English. It is a special kind of dual learning \cite{he2016dual,xia2017dualijcai,xia2017dual,li2019explicit} in both directions of language pairs. Currently, state-of-the-art UNMT models are based on a sequence-to-sequence encoder--decoder architecture using Transfomer  \cite{vaswani2017attention}, similar to supervised NMT models.

For ease of expression, in the remainder of this paper, we denote the monolingual training data space of the source $\mathcal{S}$ and target $\mathcal{T}$ languages as $\phi_\mathcal{S}$ and $\phi_\mathcal{T}$. The parallel training data space between languages $\mathcal{S}$ and $\mathcal{T}$ is represented as $\phi_{\mathcal{S}-\mathcal{T}}$. The translation direction symmetry of the UNMT model training implies that the translation direction problem $\mathcal{S} \rightarrow \mathcal{T}$ is the same as $\mathcal{T}  \rightarrow \mathcal{S}$\footnote{In UNMT, translation is bidirectional, so ``source'' and ``target'' languages only indicate translation direction for using model. Essentially, $\mathcal{S}$ and $\mathcal{T}$ are symmetrical and exchangeable.}.

In general, the NMT model with parameters $\theta_{\mathcal{S} \rightarrow \mathcal{T}}$ models the conditional probability $\mathbb{P}(\textbf{\textit{t}}|\textbf{\textit{s}})$ of the translated sequence $\textbf{\textit{t}}$. The model parameters $\theta_{\mathcal{S} \rightarrow \mathcal{T}}$ are trained to maximize the following likelihood on the parallel training data space:
\begin{equation}
\footnotesize
\mathcal{L}(\theta_{\mathcal{S} \rightarrow \mathcal{T}}) = \mathbb{E}_{\langle \textbf{\textit{s}},\textbf{\textit{t}} \rangle \sim \phi_{\mathcal{S}-\mathcal{T}}} [-\log \mathbb{P}(\textbf{\textit{t}}|\textbf{\textit{s}}; \theta_{\mathcal{S} \rightarrow \mathcal{T}})].
\end{equation}

As there is a lack of cross-lingual sentence alignment information, the current UNMT models, despite their differences in training methods and structure, reach a consensus over the use of the parallel data that was iteratively generated by the BT method.
Specifically, for a monolingual sentence of target language $\textbf{\textit{t}} \in \phi_\mathcal{T}$, a source translation $\tilde{\textbf{\textit{s}}}$ is generated using the primal $\mathcal{T}  \rightarrow \mathcal{S}$ translation model $\mathbb{P}(\cdot|\textbf{\textit{t}}, \theta_{\mathcal{T} \rightarrow \mathcal{S}})$, then $\tilde{\textbf{\textit{s}}}$ and $t$ form a pseudo-parallel pair $\langle \tilde{\textbf{\textit{s}}},\textbf{\textit{t}} \rangle$  for $\mathcal{S}  \rightarrow \mathcal{T}$  model training. Similarly, the generated pseudo-parallel pair $\langle \tilde{\textbf{\textit{t}}},\textbf{\textit{s}} \rangle$ for a monolingual sentence $\textbf{\textit{s}}$ in the source language is also used for training the $\mathcal{T}  \rightarrow \mathcal{S}$ model.

The likelihood of the reconstructions $\textbf{\textit{t}} \rightarrow \tilde{\textbf{\textit{s}}} \rightarrow \textbf{\textit{t}}$ and $\textbf{\textit{s}} \rightarrow \tilde{\textbf{\textit{t}}} \rightarrow \textbf{\textit{s}}$ for the UNMT model is maximized over the BT process  according to:
\begin{equation}
\footnotesize
\mathcal{L}(\theta_{\mathcal{S} \rightarrow \mathcal{T}})=\mathbb{E}_{\tilde{\textbf{\textit{s}}} \sim \mathbb{P}(\cdot|\textbf{\textit{t}}, \theta_{\mathcal{T} \rightarrow \mathcal{S}}),\textbf{\textit{t}} \sim \phi_{\mathcal{T}}} [-\log \mathbb{P}(\textbf{\textit{t}}|\tilde{\textbf{\textit{s}}}; \theta_{\mathcal{S} \rightarrow \mathcal{T}})],
\end{equation}
\begin{equation}
\footnotesize
\mathcal{L}(\theta_{\mathcal{T} \rightarrow \mathcal{S}})=\mathbb{E}_{\textbf{\textit{s}} \sim \phi_{\mathcal{S}},\tilde{\textbf{\textit{t}}} \sim \mathbb{P}(\cdot|\textbf{\textit{s}}, \theta_{\mathcal{S} \rightarrow \mathcal{T}})} [-\log \mathbb{P}(\textbf{\textit{s}}|\tilde{\textbf{\textit{t}}}; \theta_{\mathcal{T} \rightarrow \mathcal{S}})].
\end{equation}

Finally, the BT process is optimized by minimizing the following objective function:
\begin{equation}
\footnotesize
\mathcal{L}_{\textbf{BT}}(\mathcal{S}, \mathcal{T}) = \mathcal{L}(\theta_{\mathcal{S} \rightarrow \mathcal{T}}) + \mathcal{L}(\theta_{\mathcal{T} \rightarrow \mathcal{S}}).
\end{equation}

\section{Reference Language based UNMT}
In this section, we introduce the reference language-based UNMT framework and present our three kinds of reference agreement utilization approaches: reference agreement translation (RAT), reference agreement back-translation (RABT), and cross-lingual back-translation (XBT). These approaches are illustrated in Figure \ref{methods}.

\subsection{Framework and Reference Agreement}
Figure \ref{RUNMT}(a) demonstrates the traditional pivot translation schema in supervised NMT, subfigure \ref{RUNMT}(b) shows the multilingual UNMT, and subfigure \ref{RUNMT}(c) is our proposed reference language-based UNMT framework.
When applying pivot translation to UNMT, any language pair in UNMT can be directly trained without any parallel data, which allows translation in both directions due to the nature of UNMT. 
Thus, the traditional pivot schema ($\mathcal{S} \rightarrow \mathcal{P} \rightarrow \mathcal{T}$) is not necessary when applying pivot translation to UNMT; using a third language (usually a common language) is a more suitable practice for UNMT. 
In order to distinguish from the pivot language in traditional pivot translation, we define the language used to enhance the performance of translation $\mathcal{S} \rightarrow \mathcal{T}$ in UNMT as the reference language $\mathcal{R}$, regardless of whether the translation schema is $\mathcal{S} \rightarrow \mathcal{R} \rightarrow \mathcal{T}$ as the bridge or $\mathcal{S} \rightarrow \mathcal{T}$ directly.

In this paper, the reference agreement refers to the cross-lingual equivalence (i.e., translation agreement) provided by bilingual parallel sentence pairs between the reference language and the source or target language of the translation. 

\subsection{Reference Agreement Translation}

In the absence of supervision signals, the quality of machine translation across languages cannot be effectively evaluated. That is, a suitable cross-lingual quality evaluation function $quality(\textbf{\textit{s}}, \tilde{\textbf{\textit{t}}})$ cannot be defined in cases where only the source and target generation are provided. As a result, the quality of synthetic pseudo-parallel pairs $\langle \tilde{\textbf{\textit{s}}},\textbf{\textit{t}} \rangle$ and $\langle \tilde{\textbf{\textit{t}}}, \textbf{\textit{s}} \rangle$ in BT cannot be guaranteed, which limits the performance of UNMT.

RAT refers to the simultaneous translation of the parallel sentences of languages $\mathcal{S}$ and $\mathcal{R}$ into the target language $\mathcal{T}$. The two translations should be in agreement (i.e., the same). Therefore, this agreement in the translations from different sources can be used to collaboratively evaluate the generated quality, and it thus forms a new quality evaluation function $quality(\textbf{\textit{s}}, \textbf{\textit{r}}, \tilde{\textbf{\textit{s}}}, \tilde{\textbf{\textit{r}}})$.

Based on this premise, we propose a detailed implementation for the RAT approach, enabling reference agreement functions with BT during the UNMT training process and resulting in improved translation agreement, as shown in Figure \ref{methods}(b).
Specifically, RAT requires the two translation models to generate an agreed-upon translation by taking votes. We use this agreed-upon translation as the target and form pseudo-parallel data from the input of each language to train both of the models.

Specifically, for a parallel sentence pair $\langle \textbf{\textit{s}},\textbf{\textit{r}} \rangle$, we would ideally have $\mathbb{P}(\cdot|\textbf{\textit{s}}; \theta_{\mathcal{S} \rightarrow \mathcal{T}}) = \mathbb{P}(\cdot|\textbf{\textit{r}}; \theta_{\mathcal{R} \rightarrow \mathcal{T}})$, as stated for RAT; however, as the two models $\theta_{\mathcal{S} \rightarrow \mathcal{T}}$ and $\theta_{\mathcal{R} \rightarrow \mathcal{T}}$ are trained on different data, the agreement may be corrupted. Therefore, we combine the two models  to obtain the agreed-upon translation output $\tilde{\textbf{\textit{t}}}_a$:
\begin{equation}
\footnotesize
\tilde{\textbf{\textit{t}}}_a \sim \mathbb{P}(\cdot|\textbf{\textit{s}}, \textbf{\textit{r}}; \theta_{\mathcal{S} \rightarrow \mathcal{T}}, \theta_{\mathcal{R} \rightarrow \mathcal{T}}),
\end{equation}
where $\mathbb{P}(\cdot|\textbf{\textit{s}}, \textbf{\textit{r}}; \theta_{\mathcal{S} \rightarrow \mathcal{T}}, \theta_{\mathcal{R} \rightarrow \mathcal{T}})$ is
\begin{equation}
\footnotesize
\prod_{i=1}^{J} [ \frac{1}{2}(\mathbb{P}(\cdot|\textbf{\textit{s}}, \tilde{\textbf{\textit{t}}}_{<i}; \theta_{\mathcal{S} \rightarrow \mathcal{T}}) + \mathbb{P}(\cdot|\textbf{\textit{r}}, \tilde{\textbf{\textit{t}}}_{<i}; \theta_{\mathcal{R} \rightarrow \mathcal{T}}))],
\end{equation}
where $\tilde{t}_{< i}$ stands for tokens that have been generated prior to the i-generation step.
Finally, two synthetic sentence pairs $\langle \textbf{\textit{s}},\tilde{\textbf{\textit{t}}}_a \rangle$ and $\langle \textbf{\textit{r}},\tilde{\textbf{\textit{t}}}_a \rangle$ are used to train the models $\mathcal{S} \rightarrow \mathcal{T}$ and $\mathcal{R} \rightarrow \mathcal{T}$. Since the silver learning target is optimized, the smoothed cross-entropy loss $\mathcal{L}_\epsilon$ is used instead of the ordinary cross-entropy loss $\mathcal{L}$. The learning objective for RAT can be written as:
\begin{equation}
\footnotesize
\mathcal{L}_{\textbf{RAT}}(\mathcal{S}, \mathcal{T}, \mathcal{R}) = \mathcal{L}_\epsilon(\theta_{\mathcal{S} \rightarrow \mathcal{T}}) + \mathcal{L}_\epsilon(\theta_{\mathcal{R} \rightarrow \mathcal{T}}),
\end{equation}
where $\epsilon$ is the smoothing control value indicating the uncertainty of the target for the model.

\subsection{Reference Agreement Back-translation}

Motivated by the RAT approach, the input language sentences and agreed-upon translations form two synthetic parallel sentences. 
With these regularized pseudo-parallel sentences, we not only train the $\mathcal{S} \rightarrow \mathcal{T}$ and $\mathcal{R} \rightarrow \mathcal{T}$ forward-translation models (as the generation direction is the same as the training direction), but also train the BT models, i.e., $\mathcal{T} \rightarrow \mathcal{S}$ and $\mathcal{T} \rightarrow \mathcal{R}$. This gives the RABT training approach shown in Figure \ref{methods}(c). The learning objective of RABT can be described as:
\begin{equation}
\footnotesize
\mathcal{L}_{\textbf{RABT}}(\mathcal{S}, \mathcal{T}, \mathcal{R}) = \mathcal{L}(\theta_{\mathcal{T} \rightarrow \mathcal{S}}) + \mathcal{L}(\theta_{\mathcal{T} \rightarrow \mathcal{R}}).
\end{equation}

\subsection{Cross-lingual Back-translation}
The traditional BT analyzed in Section \ref{unmt} and illustrated in  Figure \ref{methods}(a) allows us to train a $\mathcal{T} \rightarrow \mathcal{S}$ model with the help of an $\mathcal{S} \rightarrow \mathcal{T}$ model, and vice versa; however, this mutually beneficial training is performed entirely within one language pair. Multilingual UNMT (MUNMT) \cite{sun2020knowledge} is a special case of UNMT that is capable of translating between multiple source and target languages. Although multiple language pairs are trained jointly in MUNMT, there is an obvious shortcoming for BT: translating between language pairs that do not occur together during training, i.e., lack of optimization across language pairs. Joint training across language pairs can be performed through forced high-order BT in UNMT, which takes the form $L_1 \rightarrow L_2 \rightarrow ... \rightarrow L_{O+1} \rightarrow L_1 $, where $O$ is the translation order indicating the number of bridge languages in BT. This approach may fail because decoding through multiple noisy channels ($L_i \rightarrow L_{i+1}$) accumulates latency and compounds errors, resulting in low-quality final pseudo-parallel data between $L_{O+1}$ and $L_1$.

Although this high-order BT can expose multiple language pairs for simultaneous training, it also introduces the problem of uncontrollable intermediate translation quality. Therefore, we propose XBT based on the reference agreement. This method allows BT to remain first order while training across language pairs. XBT is a new training approach for UNMT that translates language $\mathcal{S}$ to $\mathcal{T}$ and then back-translates it to $\mathcal{R}$, or from $\mathcal{R}$ to $\mathcal{T}$ and then to $\mathcal{S}$, based on the reference agreement provided by the bilingual parallel data $\phi_{\mathcal{S}-\mathcal{R}}$ between languages $\mathcal{S}$ and $\mathcal{R}$. This training approach is illustrated in Figure \ref{methods}(d). The objective function of XBT is:
\begin{equation}
\footnotesize
\mathcal{L}_{\textbf{XBT}}(\mathcal{S}, \mathcal{T}, \mathcal{R}) = \mathcal{L}(\theta_{\mathcal{T}_\mathcal{S} \rightarrow \mathcal{R}}) +  \mathcal{L}(\theta_{\mathcal{T}_\mathcal{R} \rightarrow \mathcal{S}}),
\end{equation}
where $\mathcal{T}_\mathcal{S}$ and $\mathcal{T}_\mathcal{R}$ indicate language sentences translated from $\mathcal{S}$ and $\mathcal{R}$, respectively.

\begin{table*}[t]
	\begin{center}
		\begin{small}
			\begin{tabular}{lcccccccccc}
				\toprule
				& \multicolumn{4}{c}{\textit{en}-\textit{fr}-\textit{ro}} & & \multicolumn{4}{c}{\textit{en}-\textit{zh}-\textit{ro}} &  \\
				\cmidrule{2-5} \cmidrule{7-10}
				& \multicolumn{1}{c}{\textit{en}$\rightarrow$\textit{ro}} & \multicolumn{1}{c}{\textit{ro}$\rightarrow$\textit{en}} & \multicolumn{1}{c}{\textit{fr}$\rightarrow$\textit{ro}} & \multicolumn{1}{c}{\textit{ro}$\rightarrow$\textit{fr}} & & \multicolumn{1}{c}{\textit{en}$\rightarrow$\textit{ro}} & \multicolumn{1}{c}{\textit{ro}$\rightarrow$\textit{en}} & \multicolumn{1}{c}{\textit{ro}$\rightarrow$\textit{zh}} & \multicolumn{1}{c}{\textit{zh}$\rightarrow$\textit{ro}} & \# \\
				\midrule
				PBSMT + NMT & 25.13 & 23.90 & n/a & n/a & &  25.13 & 23.90 & n/a & n/a & 1\\
				XLM & 33.30 & 31.80 & n/a & n/a & &  33.30 & 31.80 & n/a & n/a & 2\\
				MASS &  \underline{35.20} &  \underline{33.10} & n/a & n/a & &   \underline{35.20} &  \underline{33.10} & n/a & n/a & 3\\
				\midrule
				UNMT & 34.45 & 32.42 & 25.26 & 27.99 &  & 34.45 & 32.42 & 8.66 [2.31] & 10.92 [3.56] & 4 \\
				\midrule
				MUNMT & 34.44 & 32.60 & 25.31 & 27.91 &  & 33.79 & 31.82 & 8.85 [2.63] & 11.55 [3.87] & 5 \\
				\hdashline
				+ RAT & 35.83 & 33.52 & 25.66 & 28.25 &  & 34.59 & 32.12 & 9.73 [3.02] & 12.44 [3.95] & 6 \\
				+ RABT & 36.05 & 33.74 & 25.65 & 28.44 &  &  35.23 &  32.67 & 10.09 [3.30] & 12.95 [4.00] & 7 \\
				+ XBT & 36.08 &  33.84 &  \underline{25.78} &  28.45 &  & 34.76 & 32.30 &  10.54 [3.32] &  13.66 [4.03] & 8 \\
				\bf +ALL & \underline{36.14} & \underline{34.12} & 25.60 & \underline{28.89} & & \underline{35.66}  & \underline{32.88} & \underline{10.83} [3.44] & \underline{13.75} [4.24] & 9 \\	
				\midrule
				MUNMT + RNMT & 36.39 & 33.85 & 25.53 & 28.57 &  & 35.50 & 33.66 & 10.98 [3.64] & 14.42 [4.39] & 10 \\
				\hdashline
				+ RAT & 36.65 & 34.07 & 25.78 & 28.63 &  & 36.26 & 34.18 & 11.26 [3.87] & 14.77 [4.78] & 11 \\
				+ RABT & 36.84 & 34.32 & 25.75 & 29.04 &  & 36.78 & 34.26 & 11.52 [3.90] & 14.79 [5.01] & 12 \\		
				+ XBT & 37.13 & 34.66 & 26.02 & 29.11 &  & 36.31 & 34.14 & 11.80 [4.03] & 14.86 [4.98] & 13 \\
				\bf +ALL & \bf 37.27 & \bf 34.85 & \bf 26.50 & \bf 29.45  &  & \bf 37.01 & \bf 34.55 & \bf 11.92 [4.07] & \bf 15.02 [5.11] & 14 \\			
				\bottomrule
			\end{tabular}
		\end{small}
	\end{center}
	\caption{Comparison of the proposed methods with previous work (MultiBLEU). Overall best results are shown in bold (all our best results are better than the corresponding baselines at significance level $p < 0.01$ \cite{Collins:2005:CRS:1219840.1219906}). PBSMT + NMT: \cite{lample2018phrase}, XLM: \cite{conneau2019cross}, MASS: \cite{song2019mass}. In the form $x [y]$,  $x$ and $y$ respectively indicate results on in-domain and out-of-domain sets. Note, the BLEU used in \textit{ro}$\rightarrow$\textit{zh} is based on Chinese words segmented by the \textit{jieba} toolkit.}
	\label{tab:results_main}
\end{table*}

\section{Experiments and Results}

\subsection{Datasets}
We consider multilingual UNMT for four languages: English (\textit{en}), French (\textit{fr}), Romanian (\textit{ro}), and Chinese (\textit{zh}). 
To compare the impact of the relationship between the chosen reference language and the considered language pairs on the UNMT performance, we constructed two language scenarios: English--French--Romanian (\textit{en}-\textit{fr}-\textit{ro}) and English--Chinese--Romanian (\textit{en}-\textit{zh}-\textit{ro}), where English--Romanian (\textit{en}-\textit{ro}) is the main language pair considered. French and Chinese are used as the reference languages, providing the parallel corpora of English--French (\textit{en}-\textit{fr}) and English--Chinese (\textit{en}-\textit{zh}), respectively, to aid the UNMT of English--Romanian.  
English and Romanian belong to the Indo-European language family, but English belongs to the Germanic branch, whereas Romanian and French belong to the Romance branch.  French is selected to evaluate the effect of the reference language being in the homologous family. Chinese belongs to the Sino-Tibetan language family, which is a distant language from Romanian and is selected to study a different language family reference language.

For English, French, and Romanian, we used the same monolingual sentences as those extracted from the WMT News Crawl datasets for the period 2007--2017 by \citet{conneau2019cross} for a fair comparison and limited the maximum number of sentences in each language to 50 million(M), which results in 50M, 50M, and 14M sentences, respectively. For Chinese, we combined all of the sentences available in the WMT News Crawl datasets with the source sentences from the WMT'17 Chinese--English translation task, leading to 26M sentences. For the parallel data of \textit{en}-\textit{fr} and \textit{en}-\textit{zh} introduced by the two experimental settings, we only use those provided by MultiUN \cite{ziemski2016united}. Finally, the size of the resulting language pair parallel dataset is about 10M.

In both scenarios, we evaluated each language pair except for \textit{en}-\textit{fr} and \textit{en}-\textit{zh}, for which the relevant parallel data was used for reference agreement. Following previous studies, newstest 2016 was used to evaluate the \textit{en}-\textit{ro} language pair. For \textit{fr}-\textit{ro}, we sampled 5K sentence pairs from OPUS \cite{tiedemann2012parallel} for evaluation, while for \textit{zh}-\textit{ro}, we use the religious and educational parallel data for out-of-domain evaluation and collected 2K news parallel sentences for in-domain evaluation. 
In detail, as data for \textit{fr}-\textit{ro}, we used GlobalVoices\footnote{http://casmacat.eu/corpus/global-voices.html}, OpenSubtitles \cite{lison2016opensubtitles2016}, and MultiParaCrawl\footnote{http://paracrawl.eu}, whereas for \textit{zh}-\textit{ro}, Bible-uedin \cite{christodouloupoulos2015massively}, Tanzil, and the QCRI Educational Domain Corpus (QED) \cite{abdelali2014amara} were used. Because these parallel corpora between \textit{zh}-\textit{ro} are in religious and educational domains only, which are far away from the news domain of training data, we also collected a parallel corpus (2K in size) of \textit{zh}-\textit{ro} for in-domain evaluation. 

The Moses scripts \cite{koehn2017six} were used for tokenization of \textit{en}, \textit{fr}, and \textit{ro}, and the jieba toolkit\footnote{https://github.com/fxsjy/jieba} was used for word segmentation on \textit{zh}. In particular, following \citet{sennrich2016edinburgh}, we removed diacritics from \textit{ro}. For \textit{zh}, to avoid confusion between Hong Kong Standard Traditional Chinese (\textit{zh\_hk}: QED), Taiwan Standard Traditional Chinese (\textit{zh\_tw}: Bible-uedin), and Simplified Chinese (\textit{zh}: Tanzil and monolingual training data), we used opencc\footnote{https://github.com/BYVoid/OpenCC} to convert  \textit{zh\_hk} and \textit{zh\_tw} to simplified Chinese.

\begin{table}
	\renewcommand\tabcolsep{5.0pt}
	\begin{center}
		\begin{small}
			\begin{tabular}{c|cccc}
				\toprule
				& \textit{en} & \textit{fr} & \textit{ro} & \textit{zh} \\
				\midrule
				\textit{en}-\textit{ro} & 6.5 / 64.3 & - & 4.9 / 68.3 & - \\
				\textit{fr}-\textit{ro} & - &  4.1 / 68.7 & 4.9 / 68.5 & - \\
				\textit{zh}-\textit{ro} & - & - & 5.3 / 65.8 & 11.5 / 52.9 \\
				\textit{en}-\textit{fr}-\textit{ro} & 6.9 / 60.1 & 4.2 / 68.4 & 5.0 / 68.1 & - \\
				\textit{en}-\textit{zh}-\textit{ro} & 7.4 / 53.8 & - & 5.5 / 64.9 & 11.4 / 53.4 \\
				\bottomrule
			\end{tabular}
		\end{small}
	\end{center}
	\caption{Perplexity / Accuracy for masked language modeling in different languages joint pre-training.}\label{ppl}
\end{table}

\subsection{Baselines}
Our baseline models follow XLM \cite{conneau2019cross}, with the following refinements:
\paragraph{UNMT} 
\citet{lample2018unsupervised,artetxe2019massively,song2019mass} have demonstrated the importance of pre-training, which is a key ingredient of UNMT. 
\citet{conneau2019cross} used masked language modeling (MLM) to pre-train the full model for the initialization step before applying a denoising autoencoder and BT training step. Therefore, we take the XLM architecture proposed by \citet{conneau2019cross} as our backbone baseline model.

\paragraph{MUNMT} 
Our method studies the impact of adding a reference language to the existing UNMT language pair, which makes our model essentially multilingual. Therefore, MUNMT is the baseline for comparison. We adopt a multi-language joint vocabulary and training with a shared encoder and decoder for language model pre-training, denoising, and BT as the basis of our backbone, UNMT (XLM). Thus, with these settings, the MUNMT model can take advantage of multilingualism.

\paragraph{MUNMT + RNMT}
Furthermore, as we use a parallel corpus that exists between the reference language and the unsupervised translation language, for a fairer comparison, we consider adding a supervised neural machine translation between the source and reference language (RNMT) as an extra training step on the basis of MUNMT so that supervised and unsupervised training are performed jointly. This baseline is named MUNMT + RNMT.

In all our baselines, the byte pair encoding (BPE) code size is set to 60K, and the model hyper-parameters are consistent with those of XLM. The smoothing value $\epsilon$ in RAT is set to $0.1$. 

\subsection{Main Results and Analysis}\label{results_analysis}
This section examines the effectiveness of the proposed RUNMT framework\footnote{Code available at \url{https://github.com/bcmi220/runmt}.}. The main results\footnote{Notably, concurrent works \cite{liu2020multilingual,bai2020unsupervised,garcia2020multilingual} also explore the case of using auxiliary parallel data effects under the MUNMT setting, where all of these works share similarities in multilingualism motivation. Due to the inconsistency of the parallel corpora used, the results are not directly comparable, so we don't include their results in the table.} are presented in Table \ref{tab:results_main}. Row \#4 reports the replicated results of the XLM architecture \cite{conneau2019cross} based on the training of each language pair individually. Our UNMT basically reproduces XLM's results, and it also makes some improvements over the original (probably because of differences in data sampling). Thus, our approach offers a strong baseline performance. Compared with the current state-of-the-art method MASS \cite{song2019mass}, our baseline performance is slightly lower. This is because MASS adopts the new masked sequence to sequence the pre-training method, and the improvement of our method is orthogonal to the pre-training improvement.

For the MUNMT baseline, as shown in \#5, the results are basically consistent with the UNMT results we replicated in \#4, with some slight fluctuations, indicating the joint training of language pairs alone cannot make full use of multilingualism. Compared with MUNMT, MUNMT + RNMT (\#10) is a very strong method for using an otherwise irrelevant corpus through a reference language. 

As shown in Table \ref{ppl}, the performance (perplexity/accuracy) of joint pre-training on all languages is worse than that of pre-training on individual language pairs; however, for distant language pairs, adding a close reference language for joint pre-training will improve performance compared to pre-training on only the distant language pair. Therefore, in \#5 and \#10, the performance of \textit{en}-\textit{ro} in \textit{en}-\textit{fr}-\textit{ro} and \textit{en}-\textit{zh}-\textit{ro} is inconsistent in part due to pre-training. Similarly, comparing the performance of \textit{en}-\textit{ro} and \textit{zh}-\textit{ro} in UNMT and MUNMT, the performance of \textit{zh}-\textit{ro} in MUNMT is better than that in UNMT, indicating that transfer learning plays a role in joint training and the performance in \textit{en}-\textit{ro}  worsens, indicating that joint training a close language pair with a distant language will result in a decline in its UNMT results.

The three specific approaches (RAT, RABT, and XBT) of the proposed RUNMT framework have achieved performance improvements over strong baselines, showing the effectiveness of our proposed approaches. Among them, RAT and RABT both use agreed-upon translations and their inputs to form pseudo-parallel data for training the model: RAT uses the noisy synthetic data as the target, while RABT uses the noisy synthetic data as the source. The results in \#6 and \#10 show that although RAT with a smoothing mechanism can improve the baselines' performance, the improved result is weaker than RABT in \#7 and \#12, which use the golden sentences as the target. Comparing RABT and XBT, the gap in performance is relatively small. XBT has a greater average improvement (\#8 and \#13), indicating that agreement across language pairs is more effective in MUNMT than agreement within a language pair. In addition, combining the three approaches by optimizing them one by one in an update step, with the results shown in \#9 and \#14, further improved the performance, indicating that the agreement across language pairs and internal agreement within a language pair are complementary.

In Table \ref{tab:results_main}, we also report the results of different domains within \textit{zh}-\textit{ro}, where the results in-domain are significantly higher than the results out-of-domain, indicating that the domain problem is also important for UNMT. Our approaches have also obtained consistent improvements over different domains, further verifying the effectiveness of the method. 

\subsection{Comparison with Pivot Translation}

To alleviate the difficulty of lack of bilingual corpora, there are two solutions, the latest uses UNMT in an NMT framework, while the previous solution is pivot translation (usually in an SMT setting), in which the pivot language acts as a bridge creating a path from source to target languages, i.e. $\mathcal{S} \rightarrow \mathcal{P}$ and $\mathcal{P} \rightarrow \mathcal{T}$ across parallel corpora. Our proposed RUNMT is similar to pivot translation, as both seek help from a third language when there is a lack of parallel corpora between the source language and target language. The difference is that our RUNMT requires only one parallel corpus between source and reference languages, while pivot translation requires two: between source and pivot languages, and between pivot and target languages. 

In order to make a fairer comparison between the proposed RUNMT framework and the pivot translation framework, we conducted the following experiments in \textit{zh}-\textit{ro} translation: choosing \textit{en} as the reference language (in RUNMT) or the pivot language (in PT). The two frameworks are evaluated in two settings: one in which only one parallel corpus ($\textit{zh}-\textit{en}$) is provided as claimed in RUNMT, and the other in which two parallel corpora ($\textit{zh}-\textit{en}$ and $\textit{en}-\textit{ro}$) are provided as required in PT. Since adding a parallel corpus in our proposed RUNMT framework requires only adding additional training techniques without modifying the structure or training a new model, our RUNMT can also conveniently adapt to the setting of two parallel corpora. In order to adapt to the setting where only one parallel corpus is provided, the PT framework adopts the supervised NMT model that trains $\mathcal{S}$ to $\mathcal{P}$ ($\textit{zh} \rightarrow \textit{en}$ or $\textit{en} \rightarrow \textit{zh}$ in this experiment) and the UNMT model that trains $\mathcal{P}-\mathcal{T}$ ($\textit{en}-\textit{ro}$). For the $\textit{en}-\textit{zh}$ parallel corpus added in this setting, since MultiUN does not contain this pair, we use the training set provided by WMT'16.

The experimental results show that RUNMT is effective in not only the new case of only one parallel corpus provided, but also the traditional case of two parallel corpora provided, indicating that RUNMT generally makes better use of multilingualism. Additionally, it can be seen from the results that if the first pass of pivot translation is performed by a worse-performing model, error propagation will affect the overall performance, while direct translation in RUNMT will not be affected by this.

\begin{table}[t!]
	\small
	\centering
	\setlength{\tabcolsep}{4pt}
	\resizebox{\linewidth}{!}{
		\begin{tabular}{lcccc}  
			\toprule  
			\multirow{2}{*}{}
			&\multicolumn{2}{c}{$\textit{zh} \rightarrow \textit{ro}$}
			&\multicolumn{2}{c}{$\textit{ro} \rightarrow \textit{zh}$}
			\cr  
			\cmidrule(lr){2-3} 
			\cmidrule(lr){4-5} 
			& $\textit{zh} - \textit{en} \cdots  \textit{ro}$ & $\textit{zh} - \textit{en} - \textit{ro}$ & $\textit{ro} \cdots \textit{en} -  \textit{zh}$ & $\textit{ro} - \textit{en} - \textit{zh}$ \cr
			\midrule
			PT & 14.93 & 19.78 & 6.95 & 13.90 \cr 
			RUNMT & \bf 15.02 &  \bf 20.10 & \bf 11.92 & \bf 14.56 \cr
			\bottomrule  
		\end{tabular}
	}
	\caption{Comparison between RUNMT with traditional PT framework, where ``$\rightarrow$" represents the direction of translation, ``$-$" represents supervised NMT with parallel corpus, and ``$\cdots$" represents UNMT with only monolingual data.}\label{runmt_pt_results}
\end{table}

\section{Ablation}

\begin{figure}[t!]
	\setlength{\abovecaptionskip}{0pt}
	\begin{center}
		\pgfplotsset{height=5.6cm,width=8cm,compat=1.15,every axis/.append style={thick},every axis legend/.append style={
				at={(0.8,-0.27)}},legend columns=2}
		\begin{tikzpicture}
		\tikzset{every node}=[font=\small]
		\begin{axis}
		[width=8cm,enlargelimits=0.13, tick align=outside, ymin=34, ymax=37, xticklabels={ $0$, $1K$,$10K$, $100K$, $1M$,$10M$},
		xtick={0,1,2,3,4,5},
		ytick={34,35,36,37},
		ylabel={BLEU},
		ylabel style={align=left},xlabel={Sizes},font=\small]
		\addplot+ [sharp plot, densely dashed, mark=*,mark size=1.2pt,mark options={mark color=cyan}, color=cyan] coordinates
		{(0,34.44)(1,34.67)(2,34.81)(3,35.33)(4,35.97)(5,36.39)};\label{plot_x}
		\addlegendentry{\tiny MUNMT + RNMT}

		\addplot+[sharp plot, mark=diamond*,mark size=1.2pt,mark options={solid,mark color=orange}, color=orange] coordinates
		{(0,34.44)(1,35.10)(2,35.47)(3,36.12)(4,36.24)(5,36.65)};\label{plot_y}
		\addlegendentry{\tiny RAT}
		
		\end{axis}
		
		\begin{axis}
		[width=8cm,enlargelimits=0.13, tick align=outside, ymin=34, ymax=37,  xticklabels={ $0$, $1K$,$10K$, $100K$, $1M$,$10M$},
		axis y line=none,
		axis x line=none,
		xtick={0,1,2,3,4,5},
		ytick={34,35,36,37},
		ylabel={},xlabel={Sizes},font=\small]
		\addlegendimage{/pgfplots/refstyle=plot_x}\addlegendentry{\tiny MUNMT + RNMT} 
		\addlegendimage{/pgfplots/refstyle=plot_y}\addlegendentry{\tiny + RAT} 
		
		\addplot+[sharp plot, mark=square*,mark size=1.2pt,mark options={solid,mark color=red}, color=red] coordinates
		{(0,34.44)(1,35.25)(2,35.74)(3,36.52)(4,36.62)(5,36.84)};
		\addlegendentry{\tiny + RABT}
		
		\addplot+[sharp plot, mark=triangle*,mark size=1.2pt,mark options={solid,mark color=blue}, color=blue] coordinates
		{(0,34.44)(1,35.17)(2,35.52)(3,36.50)(4,36.76)(5,37.13)};
		\addlegendentry{\tiny + XBT}
		
		\end{axis}
		\end{tikzpicture}
		\caption{\label{fig:data} Performances of different parallel data sizes for MUNMT + RNMT $\textit{en} \rightarrow \textit{ro}$ with RUNMT.}
		
	\end{center}
\end{figure}
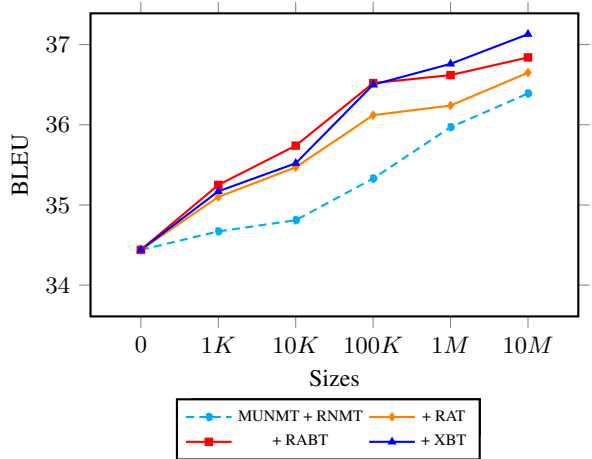

\paragraph{Effects of Parallel Data Scale}
In order to analyze the influence of the scale of reference and source language parallel data on the performance of MUNMT and our proposed approaches, we compared the performance of $\textit{en} \rightarrow \textit{ro}$ on five different parallel corpus sizes: $1K, 10K, 100K, 1M, 10M$ together with UNMT baseline, and the results are shown in Figure \ref{fig:data}. 

It shows that although MUNMT + RNMT has been a very strong method for using an otherwise irrelevant corpus through a reference language compared to MUNMT, our proposed RUNMT framework can still improve on various parallel data scales, which verifies the generalization of our method. 
In addition, in the setting with low parallel data, RABT shows a better growth effect than XBT, and when the parallel resources reach a certain scale, XBT surpasses RABT, indicating that agreement training for cross-language pairs requires more parallel data than does agreement within a language pair.
Furthermore, the effect of back-translation enhancement in all cases is better than that of forward-translation, which shows that the golden target is better than the silver target in UNMT. Finally, in low-resource settings, our methods have achieved a greater relative improvement, indicating that our methods mine the information of partially relevant parallel data to a greater extent for enhancing UNMT.

\paragraph{Analysis of Intermediate Translation Quality in BT}\label{quality}

To verify the problem of uncontrollable intermediate quality in the back-translation, we perform experiments on the distant language pair \textit{zh}-\textit{ro} and report the results of translation direction $\textit{ro} \rightarrow \textit{zh}$. The reason for choosing \textit{zh}-\textit{ro} is that Chinese and Romanian characters can be directly distinguished by using unicode encoding. We define BT-BLEU as the BLEU of $\textbf{\textit{s}} \in \mathcal{S}$ and $\tilde{\textbf{\textit{s}}}$ generated in the $\mathcal{S} \rightarrow \mathcal{T} \rightarrow \mathcal{S}$ back-translation process, and we introduce this metric in the evaluation phase. We calculate the ratio of the generated Chinese token (subword) to the total number of generated tokens to reflect the intermediate quality of the back-translation from the side. The experimental results are shown in Table \ref{bt_q}.

\begin{table}[t!]
	\begin{center}
		\begin{small}
			\begin{tabular}{l|ccc}
				\toprule
				& BLEU & BT-BLEU & RATIO \\
				\midrule
				UNMT & 10.92 & 31.47 & 2.17\% \\
				\midrule
				MUNMT & 11.55 &  31.98 & 2.01\% \\
				MUNMT + RABT & 12.95 & 32.52 & 1.69 \% \\
				MUNMT + XBT & 13.66 & 33.16 & 1.62 \% \\
				\bottomrule
			\end{tabular}
		\end{small}
	\end{center}
	\caption{Intermediate Translation Quality in BT.}\label{bt_q}
\end{table}

The results show that the growth trend of BLEU is consistent with the downward trend of the ratio of Chinese tokens in the Romanian translations, which has a notable correlation, indicating that this ratio can indeed reflect the training effect of the model to a certain extent.

In addition, compared with MUNMT, our methods improve the quality of intermediate translations, bring BT-BLEU improvement, and reduce the proportion of Chinese tokens in Romanian translations, thus verifying the effectiveness of our methods.

\section{Related Work}

With the development of the deep neural network \cite{he-etal-2018-syntax,cai-etal-2018-full,li-etal-2018-unified,zhou-zhao-2019-head,zhang2019effective,zhang2019semantics,zhou2019parsing}, UNMT \cite{artetxe2017learning,lample2018unsupervised,lample2018phrase,conneau2019cross,song2019mass} has attracted widespread attention in academic research, as only large-scale monolingual corpora are required for training. 
The performance of UNMT has benefited from language model pre-training, denoising autoencoders, and BT techniques between similar languages such as English and French, but still lags behind that of supervised NMT for distant languages such as Chinese and English. 
\citet{conneau2019cross} extended the generative language model pre-training approach to multiple languages and showed that cross-lingual pre-training could be effective for MUNMT. 
Aside from the convenience of translation among multiple language pairs, including unseen language pairs, transfer learning should be considered when low-resource languages are trained together with rich-resource ones.
As discussed by \citet{arivazhagan2019missing}, MUNMT usually performs worse than pivot-based supervised NMT; however, the pivot-based method easily experiences a computationally expensive quadratic growth in the number of source languages and suffers from the error propagation problem. 
\citet{arivazhagan2019missing} addressed the zero-shot generalization problem that some translation directions have not been optimized well due to a lack of parallel data. 
\citet{al-shedivat-parikh-2019-consistency} introduced a consistent agreement-based training method that encourages the model to
produce equivalent translations of parallel sentences in zero-shot translation, which share similarities with our RAT approach. 
However, in terms of a specific implementation, because of the differences between UNMT and NMT, we have provided three new UNMT methods, and have alleviated the problem of uncontrollable intermediate BT quality in UNMT. \citet{arivazhagan2019missing} addressed the issue of transfer learning between language pairs with parallel data where there is a lack of parallel corpora in multilingual supervised NMT.
As for the agreement in UNMT, \cite{sun2019unsupervised} investigate the  enhancement of unsupervised bilingual word embedding agreement in the UNMT training.
\citet{leng2019unsupervised} propose a multi-hop UNMT that automatically selects a good translation path for a distant language pair during UNMT. 
\citet{baijun2019cross} proposed a cross-lingual pre-training approach that makes use of the source--pivot data to pre-train the language model. 

As for the multilingualism, \citet{liu2020multilingual} proposes a multilingual denoising pre-training technique to improve machine translation tasks. \citet{bai2020unsupervised} and \citet{garcia2020multilingual} both studied the agreement across language pairs. Their method is much the same as one of our proposed approaches, XBT, which relies on the supervision signals from a parallel corpus to build a bridge between language pairs in MUNMT. Compared with these two concurrent works, the other two settings of our proposed approaches, RAT and RABT, which use the internal agreement within language pair to improve the translations, can be used not only for MUNMT, but also for semi-supervised NMT to enhance the effect of the only two languages.

\section{Conclusion}

In this work, we capitalize on the supervised NMT and UNMT use of the pivot language in pivot translation. We propose the reference language-based UNMT framework, in which a reference agreement mechanism is introduced in several implementations to better leverage the reference agreement in parallel data brought by the reference language to reduce the uncontrollable intermediate quality problem in back-translation. 
The experimental results show that we achieved an improvement over our strong baseline, and our proposed RUNMT framework is compatible with and exceeds the traditional pivot translation framework.

\bibliographystyle{acl_natbib}
\bibliography{emnlp2020}

\clearpage
\appendix

\section{Appendices}

\subsection{Pivot Translation}\label{pivot_translation}

Recent state-of-the-art NMT models are heavily dependent on a large number of bilingual language resources. Large-sized bilingual text datasets are usually readily available between common and other languages; however, for language pairs that are used less frequently, few or no bilingual resources may be available. 

Pivot translation was proposed to overcome the resource limitations for certain language pairs.
Recent state-of-the-art NMT models heavily depend on a large number of bilingual language resources. Large-sized bilingual text data sets are usually readily available between the lingua francas and other languages. However, for less-frequently used language pairs, only a limited amount or even none of the bilingual resources are available. Pivot translation was proposed to overcome resource limitations for certain language pairs due to the lack of bilingual language resources.
Instead of a direct translation between two languages for which few or no bilingual resources are available, the pivot translation approach makes use of a third language (namely the pivot language). This third language is more appropriate because of the availability of more bilingual corpora and/or its relatedness to either the source or the target language.

Pivot translation has long been studied in statistical machine translation \cite{wu2007pivot,utiyama2007comparison,paul2009importance}, supervised NMT \cite{cheng2017joint,liu2018pivot,kim2019pivot}, and UNMT \cite{leng2019unsupervised} as a means of improving the performance of low/zero-resource translations.

Formally, for the translation from language $\mathcal{S}$ to $\mathcal{T}$, the chosen pivot language is denoted as $\mathcal{P}$. The translation schema can be described as follows:
\begin{equation}
\footnotesize
\mathcal{S} \rightarrow \mathcal{P}_1 \rightarrow ... \rightarrow \mathcal{P}_K \rightarrow \mathcal{T},
\end{equation}
where $K$ is the number of pivot languages.

Recently, the development of UNMT seems to have lessened the importance of pivot translation. UNMT no longer requires bilingual parallel data between two languages, so the low/zero-resource translation problem for less-common language pairs is partially solved; however, the performance of UNMT between some distant languages in different language groups or families is still not promising, which leads researchers to reconsider pivot translation based on UNMT.

\begin{table}[t]
	\begin{center}
		\begin{small}
			\begin{tabular}{lcccc}
				\toprule
				& \multicolumn{4}{c}{\textit{en}-\textit{fr}-\textit{ro}}  \\
				\cmidrule{2-5} 
				& \multicolumn{1}{c}{\textit{en}$\rightarrow$\textit{ro}} & \multicolumn{1}{c}{\textit{ro}$\rightarrow$\textit{en}} & \multicolumn{1}{c}{\textit{fr}$\rightarrow$\textit{ro}} & \multicolumn{1}{c}{\textit{ro}$\rightarrow$\textit{fr}} \\
				UNMT & 34.45 & 32.42 & 25.26 & 27.99  \\
				\midrule
				MUNMT & 34.44 & 32.60 & 25.31 & 27.91\\
				\hdashline
				+ RAT-D & 34.71 & 33.01 & 25.42 & 28.04  \\
				+ RAT-ID & 35.83 & 33.52 & 25.66 & 28.25  \\
				\midrule
				MUNMT + RNMT & 36.39 & 33.85 & 25.53 & 28.57  \\
				\hdashline
				+ RAT-D & 36.43 & 34.55 & 25.50 & 28.59 \\
				+ RAT-ID & 36.65 & 34.07 & 25.78 & 28.63  \\			
				\bottomrule
			\end{tabular}
		\end{small}
	\end{center}
	\caption{Comparison of the proposed different RAT implementations.}
	\label{tab:results_rat}
\end{table}

\subsection{RAT-D and RAT-ID}

In this paper, the RAT method is proposed to seek the consistency of the outputs of the two translation directions, $\mathcal{S} \rightarrow \mathcal{T}$ and $\mathcal{R} \rightarrow \mathcal{T}$, when their input is parallel. In addition to the implementation described in this paper, the output distribution of $\mathcal{S} \rightarrow \mathcal{T}$ and $\mathcal{R} \rightarrow \mathcal{T}$ can also be directly computed as the agreement loss between $\mathcal{S} \rightarrow \mathcal{T}$ and $\mathcal{R} \rightarrow \mathcal{T}$. For convenience, we call this implementation RAT-D, and we call the implementation described in this paper RAT-ID.

As the two translations $\tilde{\textbf{\textit{t}}}_\mathcal{S}$ and $\tilde{\textbf{\textit{t}}}_\mathcal{R}$ from the parallel sentence pair   $\langle \textbf{\textit{s}},\textbf{\textit{r}} \rangle$ should be the same, it is clear that their probability distributions $\tilde{\textbf{\textit{d}}}_{\mathcal{S}} = \mathbb{P}(\cdot|\textbf{\textit{s}}; \theta_{\mathcal{S} \rightarrow \mathcal{T}})$ and $\tilde{\textbf{\textit{d}}}_{\mathcal{R}} = \mathbb{P}(\cdot|\textbf{\textit{r}}; \theta_{\mathcal{R} \rightarrow \mathcal{T}})$ should ideally be consistent. We would like to minimize the distance of $\tilde{\textbf{\textit{d}}}_{\mathcal{S}}$ and $\tilde{\textbf{\textit{d}}}_{\mathcal{R}}$ so that the agreement is learned by the model. The Jensen--Shannon divergence (JSD) \cite{fuglede2004jensen} is then used to compute the difference in the two distributions as the loss for RAT-D training. This is a symmetrized and smoothed version of the Kullback--Leibler divergence (KLD):
\begin{equation}\label{jsd}
\footnotesize
\begin{split}
\mathcal{L}_{\textbf{RAT-D}}(\mathcal{S}, \mathcal{T}, \mathcal{R}) & = \textbf{JSD}(\tilde{\textbf{\textit{d}}}_{\mathcal{S}} || \tilde{\textbf{\textit{d}}}_{\mathcal{R}}) = \\
\frac{1}{2}( \textbf{KLD}&(\tilde{\textbf{\textit{d}}}_{\mathcal{S}} || \textbf{\textit{M}}) + \textbf{KLD}(\tilde{\textbf{\textit{d}}}_{\mathcal{R}} || \textbf{\textit{M}})),
\end{split}
\end{equation}
where $\textbf{\textit{M}} = \frac{1}{2}(\tilde{\textbf{\textit{d}}}_{\mathcal{S}} + \tilde{\textbf{\textit{d}}}_{\mathcal{R}})$, and the KLD of distribution $Q$ from $P$ is defined as:
\begin{equation}
\footnotesize
\textbf{KLD}(P || Q)  = \sum_{i} P_i \log(\frac{P_i}{Q_i}).
\end{equation}

Autoregressive NMT models generate translations from left-to-right and stop when an \textbf{EOS} token is generated or the generation exceeds the maximum length. This leads to some length inconsistency between the two translation sequences and makes the distributions incompatible for Equation \ref{jsd}. Therefore, in the training phase, we force the translation model to generate a sequence of length $J$, which is determined as follows:
\begin{equation}
\footnotesize
J =\frac{1}{2}( (\alpha J_\mathcal{S} + \beta) + (\alpha J_\mathcal{R} + \beta)),
\end{equation}
where $J_\mathcal{S}$ and $J_\mathcal{R}$ are the lengths of the source language and reference language sentences, respectively; we set $\alpha = 1.3$ and $\beta = 5$ following previous work \cite{conneau2019cross}.

\paragraph{Differences} RAT-D and RAT-ID are the same in principle; both attempt to move two independent output distributions closer to the (weighted) average distribution through the agreement mechanism. The difference is that RAT-D is applied to the two output distributions directly; the two models are required to generate fixed-length distributions before calculating the loss, and there is no interaction between the models in the generation process. The latter point causes an error propagation problem, whereby different errors made in the two translation processes make the context in each translation increasingly different, resulting in two distributions that differ significantly. RAT-ID addresses this issue by obtaining an agreed-upon output prediction at each step, which ensures the context remains consistent in the two model generation processes.

It is shown that the effect of RAT-D is not significant compared to that of RAT-ID, which validates our belief that error propagation caused inconsistent context in the generation we analyzed.

\end{document}